\documentclass{article}

\usepackage{microtype}
\usepackage{graphicx}
\usepackage{subcaption}
\usepackage{booktabs} 
\usepackage{comment}
\usepackage{tcolorbox}
\usepackage{multirow}
\usepackage{tikz}
\newcommand{\cnum}[1]{%
  \tikz[baseline=(X.base)] \node[draw,circle,inner sep=1pt](X){#1};%
}

\usepackage{hyperref}

\usepackage[preprint]{icml2026}

\usepackage{amsmath}
\usepackage{amssymb}
\usepackage{mathtools}
\usepackage{amsthm}
\usepackage{etoc}
\usepackage{titlesec}
\usepackage{listings}
\usepackage{xcolor}
\definecolor{bg}{rgb}{0.95,0.95,0.95}

\usepackage[capitalize,noabbrev]{cleveref}

\theoremstyle{plain}

\theoremstyle{definition}

\theoremstyle{remark}

\usepackage[textsize=tiny]{todonotes}

\icmltitlerunning{Explaining Grokking in Transformers through the Lens of Inductive Bias}

\begin{document}

\twocolumn[
  \icmltitle{Explaining Grokking in Transformers through the Lens of Inductive Bias}



  \icmlsetsymbol{equal}{*}

  \begin{icmlauthorlist}
    \icmlauthor{Jaisidh Singh}{uni,mpi,taic,eliza}
    \icmlauthor{Diganta Misra}{mpi,taic,ellis}
    \icmlauthor{Antonio Orvieto}{mpi,taic,ellis}
  \end{icmlauthorlist}
  \icmlaffiliation{uni}{University of Tübingen}
  \icmlaffiliation{ellis}{ELLIS Institute Tübingen}
  \icmlaffiliation{mpi}{MPI-IS Tübingen}
  \icmlaffiliation{taic}{Tübingen AI Center}
  \icmlaffiliation{eliza}{Zuse School ELIZA}
  \icmlcorrespondingauthor{Jaisidh Singh}{\url{jaisidh.singh@student.uni-tuebingen.de}}
  \icmlkeywords{grokking,layernorm}
  \vskip 0.3in
]



\printAffiliationsAndNotice{}  
%
\begin{abstract} 
    We investigate grokking in transformers through the lens of inductive bias: dispositions arising from architecture or optimization that let the network prefer one solution over another. We first show that architectural choices such as the position of Layer Normalization (LN) strongly modulates grokking speed. This modulation is explained by isolating how LN on specific pathways shapes shortcut-learning and attention entropy. Subsequently, we study how different optimization settings modulate grokking, inducing distinct interpretations of previously proposed controls such as readout scale. Particularly, we find that using readout scale as a control for lazy training can be confounded by learning rate and weight decay in our setting. Accordingly, we show that features evolve continuously throughout training, suggesting grokking in transformers can be more nuanced than a lazy-to-rich transition of the learning regime. Finally, we show how generalization predictably emerges with feature compressibility in grokking, across different modulators of inductive bias. Our code is released at \texttt{\url{https://tinyurl.com/y52u3cad}}.
\end{abstract}
\section{Introduction}
\label{sec:intro}

One of the central explanations for the success of deep learning is the strong generalization enabled by learning problem-specific features, as demonstrated across domains such as image classification~\cite{alexnet,resnet}. At the same time, works such as~\cite{chizat2019lazy} show that neural networks can perform fit the training data well even when their parameters and features change minimally during training. This apparent tension has motivated extensive study of the relationship between optimization, generalization, and feature learning, commonly framed through the distinction between \emph{lazy} and \emph{rich} regimes.
\begin{figure*}[t]
    \centering
    \includegraphics[width=0.95\linewidth]{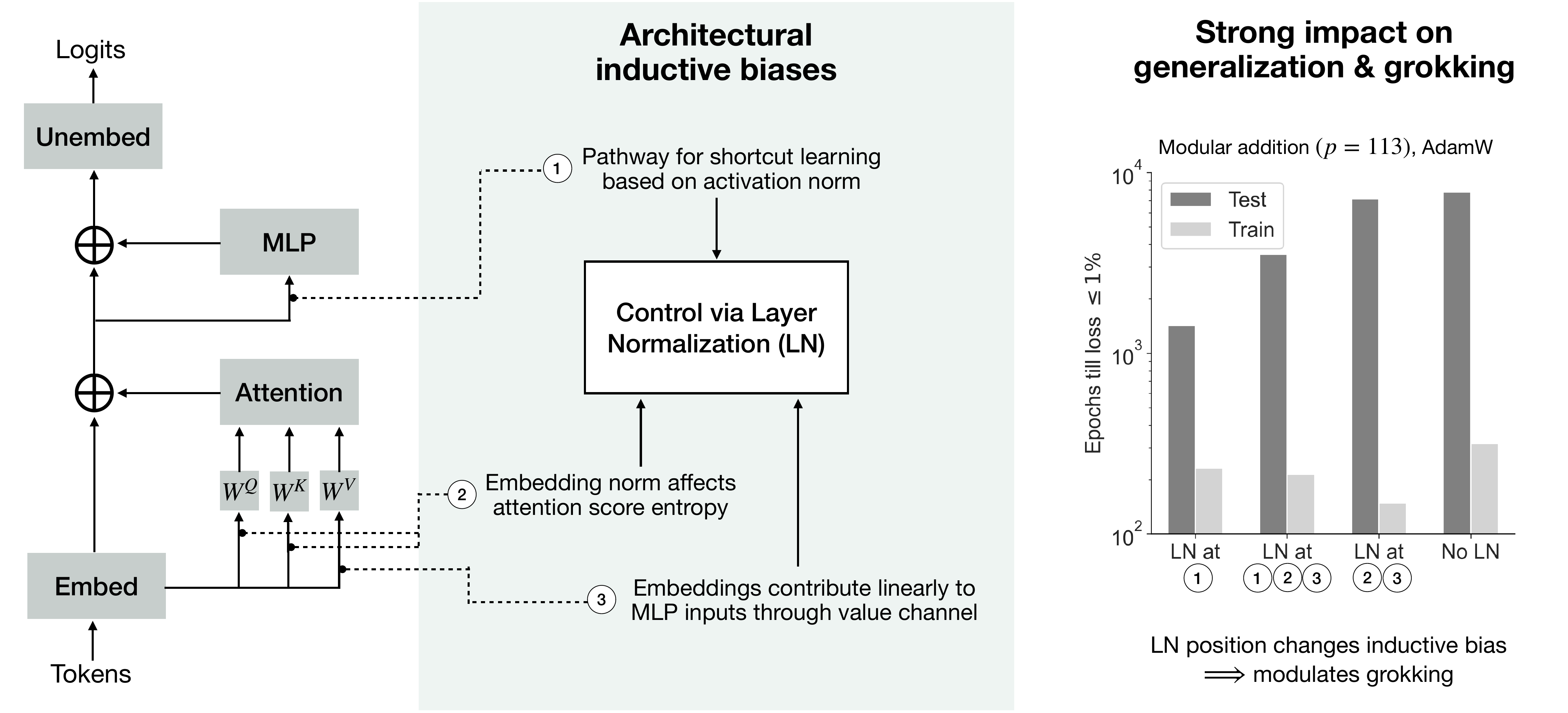}
    \caption{Different pathways express different biases in one-layer transformers. Applying layer normalization (LN) at specific positions changes the inductive bias of the network and consequently, grokking behavior.}
    \label{fig:overview}
    \vspace{-5mm}
\end{figure*}
Prior works have identified factors like read-out scale, initialization, and network width to influence lazy training, and hence tendency to suboptimally generalize~\cite{chizat2019lazy}. Oppositely, the network can operate in the rich regime\footnote{Rich regime is a term that is interchangeable with feature learning regime}, which can by admitted by conditions typically associated with lazy learning towards better performance~\cite{tp4,damian2022neural,karp2021local,shi2022theoretical}. This lazy-versus-rich characterization has become a popular lens for interpreting non-monotonic learning phenomena, including double descent~\cite{dd1,deepdd,mei2022generalization} and grokking~\cite{power2022grokking,liu2022towards,nanda2023}.

We focus on \emph{grokking}: where a network attains near-zero training loss yet only generalizes after a long delay~\cite{power2022grokking,nanda2023}. Prior work interprets grokking as a transition from lazy to rich feature learning~\cite{lazy2rich}, while complementary studies suggest that generalization arises from the emergence of structured representations shaped by hyper-parameters and data starvation~\cite{liu2022towards}. However, these perspectives do not explain why some networks discover such structure quickly while others require delays. Moreover, variation in experimental settings leads to differing interpretations~\cite{lazy2rich,liu2022towards,gromov2023grokking}.

We argue that this gap is crucially addressed through the lens of \emph{inductive bias}: disposition of a learner to prefer one solution or interpretation over another~\cite{mitchell1980need}. Naturally, inductive bias shapes learning, i.e., searching for solutions that can best explain the general rule of the data distribution~\cite{goodman1965new}. Different architectural and optimization choices accordingly afford varying preferences in the network, biasing the learning trajectory differently towards generalization. Hence, our work asks
\begin{tcolorbox}[boxrule=0.5pt, colback=gray!3]
\begin{center}
\emph{{How do inductive biases in the transformer's learning process control the rate of grokking?}} 
\end{center}
\end{tcolorbox}
To study this question, we analyze grokking through \emph{modulators}: architectural and optimization choices that systematically alter grokking behavior. Particularly, our study focuses on transformers~\cite{vaswani2017attention} learning modular addition under cross-entropy loss and adaptive optimization. This deliberately specific scope is chosen in order to avoid conflating interpretations of grokking with effects specific to optimizer, architecture, or task. 
%
%
Accordingly, we investigate the impact of inductive biases in grokking as follows.
\begin{enumerate}
    \item \emph{Inductive biases expressed through architecture modulate grokking.} First, we study architecture as a source of inductive bias. As an architectural probe, we use Layer Normalization (LN)~\cite{ba2016layernormalization}, a standard component that directly controls scale and variance propagation~\cite{onlayernorm}. Section~\ref{sec:ln_grok} shows how the impact of LN position on distinct inductive biases leads to dramatically different grokking speeds while converging to periodically structured solutions such as in~\cite{nanda2023,clock2023}.
    \item \emph{Beyond architecture, optimization choices affect inductive bias to modulate grokking.} Complementarily, we study optimization-driven inductive biases through various of choices learning rate and regularization in Section~\ref{sec:lazy}. Here, we carefully consider the assumptions required to apply previous approaches of controlling grokking, such as using readout scale as a proxy for laziness~\cite{lazy2rich}. Relating the effects of readout scale to learning rate, weight decay, and softmax-temperature, we demonstrate how optimization choices in our settings can confound characterizations of feature learning, undermining their use as reliable controls for grokking in this setting.
    \item \emph{Features move toward unified properties across different modulators and inductive biases.} Section~\ref{sec:compressibility} builds on the results of Section~\ref{sec:lazy} by showing that features evolve continuously throughout the learning process, and that generalization emerges predictably with changes in feature properties. Additionally, the evolution of features is unified across different modulators of grokking. These observations demonstrate that the network primarily operates in the rich regime. Moreover, the quality of learned features is predictably governed by the kind of inductive biases afforded by architectural and optimization choices.
\end{enumerate}
%
%
\section{Understanding How Architecture Affects Inductive Biases for Grokking}
\label{sec:ln_grok}
\begin{figure*}[t]
    \centering
    \includegraphics[width=\linewidth]{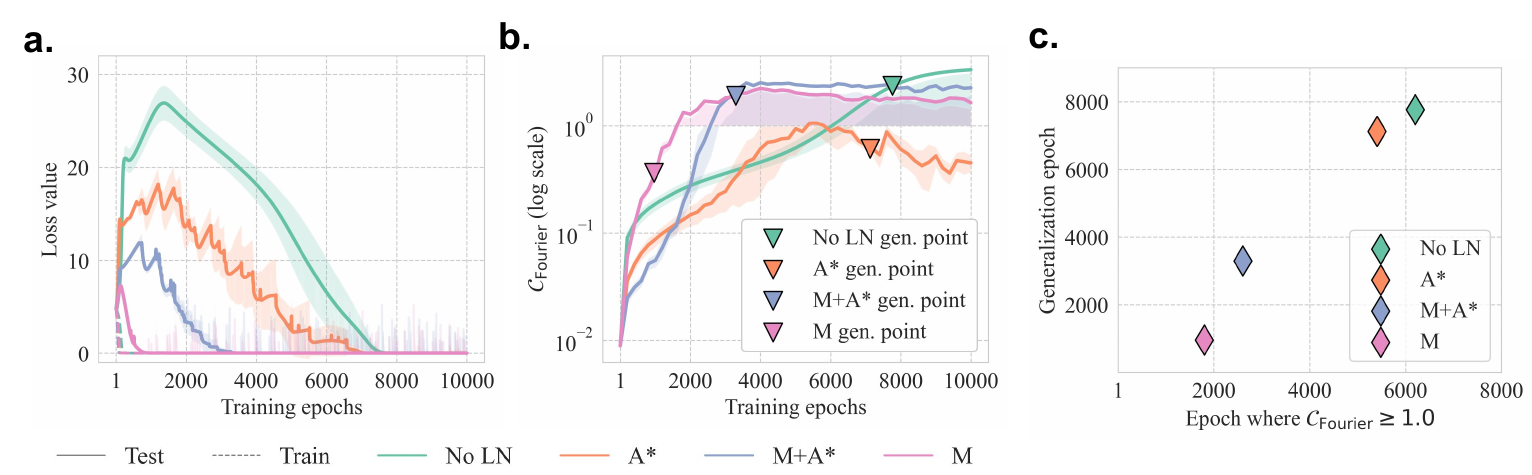}
    \caption{\textbf{(a)} Loss curves for various LN positions averaged across $5$ seeds respectively. \emph{MLP Pre-LN} and \emph{Pre} exhibit slingshots which we plot with lower   opacity. \textbf{(b)} Evolution of $\mathcal{C}_{\text{Fourier}}$ averaged across $5$ seeds along with the points where each configuration generalizes (test loss than $0.01$). \textbf{(c)} LN positions that generalize earlier show faster emergence of periodic Fourier-based structure in $\widetilde{W}_E$.}
    \label{fig:ln_main}
\end{figure*}
Our work is motivated by the large body of work done on the effect of inductive bias on generalization~\cite{barth2023howdoesthe}, which has popularly studied the network's architecture as a source of inductive bias~\cite{indbiasrazvan,boopathy2024towards,bencomo2025teasing,pmlr-v199-gowda22a}.  This lens becomes instrumental in interpreting grokking, as inductive bias directly shapes the constraints under which the learning process occurs. Thus, in this section, we ask
\begin{tcolorbox}[before skip=8pt, after skip=0pt, boxrule=0.5pt, colback=gray!3]
\begin{center}
\emph{How can inductive biases expressed through\\architecture control rate of grokking?}
\end{center}
\end{tcolorbox}
\subsection{Task \& Setting}
We use the setting of one-layer transformers learning modular addition,  it is the most widely studied setting for grokking in transformers~\cite{nanda2023,clock2023}. Given $3$ tokens $(a, b, =)$, the transformer must learn to compute the operation $(a+b)\mod p$ where $p$ is fixed and $0 \leq  a, b \leq p-1$. Next, we control the inductive biases of the transformer by using layer normalization (LN). LN is particularly chosen as the core modifier as (i) it alters learning but not model capacity, and (ii) it has been extensively 
studied~\cite{onlayernorm} in transformers. Additionally, effects of LN have crucially informed architecture design in transformers~\cite{sun2025curse,kim2025peri,karagodin2025normalization}. Complementarily, LN has been largely unexplored in grokking transformers. This allows us to use LN towards toy models of understanding the role of inductive bias in generalization via grokking.
%
%
%
\subsection{Creating Transformer Variants via LN Position}
\label{sec:ln_variants_info}
We first describe the transformer architecture by writing a forward pass given input tokens $t \in \mathbb{R}^{3 \times (p+1)}$, $p+1$ is the vocabulary size of the transformer with $p=113$. A single token at index $i \in \{1,\dots,3\}$ is denoted by $t_i$. Only the last token is unembedded as the logit. Our notation is fully described in {Table~\ref{tab:notation_table}} of {Appendix~\ref{sec:notation}}.
\begin{align}
    x_i &= W_E \ t_i + \rho_i \\
    \widetilde{x} &= \operatorname{MHSA}(x, x, x) \\ & =   \operatorname{qkv\_prod}(xW^Q, xW^K, xW^V)) \ W^O\\
    y &= x + \widetilde{x}\\
    \widetilde{y}_i &= \operatorname{MLP}(y_i) = W_2 \ \sigma(W_1 \ y_i)\\
    o &= W_U \ (y_3 + \widetilde{y}_3)
\end{align}
We define the LN operation on an input $z \in \mathbb{R}^{d}$ as 
%
    $\operatorname{LN}({z}) = \frac{{z} - \mathbb{E}[{z}]}{\sqrt{\operatorname{Var}[{z}] + \varepsilon}} \cdot \gamma + \beta$
%
($\gamma, \beta$ are learnable parameters, $\varepsilon > 0$ is a fixed scalar), and create transformer configurations with LN at different positions as given in Table~\ref{tab:ln_tab}.
%
%
\begin{table}[h]
    \centering
    \begin{tabular}{l|l}
        \toprule
        \textbf{Name} & \textbf{LN configuration} \\
        \midrule
        \textbf{No LN} & default architecture without any LN \\
        \midrule
        \multirow{2}{*}{$\mathbf{A^{*}}$} & LN only on inputs to the attention \\
         & $\widetilde{x} = \operatorname{MHSA}(\operatorname{LN}(x), \text{LN}(x), \text{LN}(x)).$ \\
        \midrule
        \multirow{3}{*}{$\mathbf{M}$+$\mathbf{A^{*}}$} & LN on inputs to both attention and MLP\\
         & $\widetilde{x} = \operatorname{MHSA}(\operatorname{LN}(x), \text{LN}(x), \text{LN}(x))$\\
         & $\widetilde{y}_i = \text{MLP}(\operatorname{LN}(y_i)).$\\
        \midrule
        \multirow{2}{*}{$\mathbf{M}$} & LN only on inputs to the MLP\\
         & $\widetilde{y}_i = \text{MLP}(\text{LN}(y_i)).$\\
        \bottomrule
    \end{tabular}
    \vspace{4pt}
    \caption{We define different configurations of LN in a one-layer transformer above. Superscript $\mathbf{*}$ in $\mathbf{A^{*}}$ denotes that inputs to each of the query, key, and value channels in $\operatorname{MHSA}(\bullet)$ are normalized.}
    \label{tab:ln_tab}
\end{table}
\subsection{Grokking under LN}
\label{sec:grokking_under_ln}
The following experiments study how position of LN within the transformer affects grokking behavior. We follow the experimental setup used in~\cite{nanda2023} supplemented by our various LN configurations defined above. Full details of the setup are given in Appendix~\ref{sec:setup}. Note that the following section uses LN as defined in Section~\ref{sec:ln_variants_info}. However, we also experiment with RMSNorm~\cite{rmsnorm} to find very similar results, given in Appendix~\ref{sec:rmsnorm}. 
%
%
\paragraph{LN position modulates grokking.} Grokking is found to be modulated by the position of LN within the transformer as shown in {Figure~\ref{fig:ln_main}}. {Figure~\ref{fig:ln_main}a} depicts that using LN in the network reduces grokking, and that certain LN configurations show less grokking than others. Particularly, $\mathbf{M}$ exhibits the fastest generalization, followed by $\mathbf{M}$+$\mathbf{A}^*$, and then finally $\mathbf{A}^*$ that generalizes only slightly faster than \textbf{No LN}. Notably, test data fit is not correlated with how fast the network learns to fit the training data, as shown in Figure~\ref{fig:overview}.
%
%
\begin{figure*}[t]
    \centering
    \includegraphics[width=\linewidth]{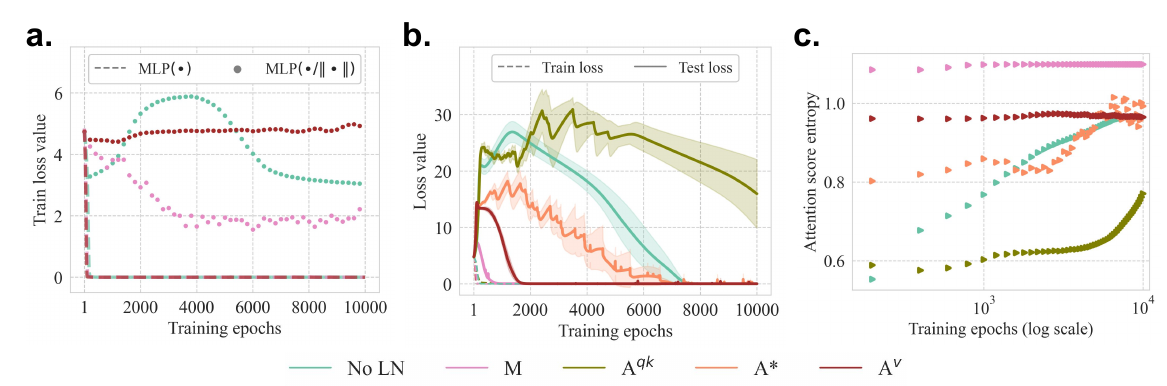}
    \caption{\textbf{(a)} The \textbf{No LN} configuration is sensitive to the norm of MLP inputs in the train loss, shown by the train loss incurred when we perturb the MLP forward pass from $\operatorname{MLP}(\bullet)$ (dashed lines) to $\operatorname{MLP}(\bullet / \|\bullet\|)$ (dots). Notably, SR-train reduces as when the network groks, showing the importance of feature scale for generalization. \textbf{(b)} LN on the inputs to the value channel of $\operatorname{MHSA}(\bullet)$ leads to fast grokking, whereas LN on query and key inputs actively hurts the network's ability to generalize. \textbf{(c)} LN on query and key inputs reduces the entropy of attention scores by removing embedding scale that acts as a degree of freedom for the network.}
    \label{fig:ib}
\end{figure*}
\paragraph{Transformers with LN learn Fourier-based solutions.} Now, we see if the network learns solutions similar to those discovered in~\cite{nanda2023,clock2023}. To do this, we Fourier transform the embedding look-up matrix $W_E$ along the vocabulary dimension after removing the $(p+1)^{\text{th}}$ embedding (for the ``='' token). This transformed matrix is denoted by $\widetilde{W}_E$. As the network approaches general solutions, $\widetilde{W}_E$ would reduce in noise, giving way to clear bands depicting Fourier modes. Since the rows of $\widetilde{W}_E$ depict Fourier modes, the ratio of between-mode variance and within-mode variance would increase as the periodic structure strengthens. We denote this ratio, which is very analogous to the F-test statistic in analysis of variance (ANOVA), as Fourier mode concentration $\mathcal{C}_{\text{Fourier}}$ in our setting.
%
%
%
%
Measuring $\mathcal{C}_{\text{Fourier}}$, we find that the relationship between how much a configuration groks with how its early $\mathcal{C}_{\text{Fourier}}$ reaches values signifying clear period structure ($\geq 1.0$) is quite symmetric. As can be observed for the average across $5$ seeds in {Figure~\ref{fig:ln_main}b}, generalization (test loss reducing beyond $0.01$) in $\mathbf{M}$ shows the earliest ascent beyond $1.0$ in terms of $\mathcal{C}_{\text{Fourier}}$, followed by $\mathbf{M}$+$\mathbf{A^{*}}$, then $\mathbf{A^{*}}$, and finally \textbf{No LN}. Furthermore, all variants arrive at periodically structured general solutions regardless of LN position. {Figure~\ref{fig:ln_main}c} shows the evolution of $\mathcal{C}_{\text{Fourier}}$ across training along with points where test loss reduces below $0.01$ as correspondence between {Figure~\ref{fig:ln_main}a}. This implies that all configurations lead to clearly emerging periodic structures in the transformer embeddings. In turn, this shows that solutions learned by transformers with LN are similar to those found in~\cite{nanda2023,clock2023}.
\subsection{How does LN position affect inductive bias?}
\label{sec:ln_ib}
%
We now turn our attention to why the previously seen grokking patterns occur for different LN positions. Below, we now isolate properties that LN positions importantly control so that the network prefers specific solutions or patterns. This is done by performing interventions that reveal LN position impacts three crucial biases for learning, that are concisely depicted in {Figure~\ref{fig:overview}}.
\paragraph{\cnum{1} $ \ $ MLP input scale.} With MLPs shown to be predominantly memorization-oriented in transformers~\cite{mlpmem}, we validate whether the MLP in our transformer can learn shortcuts by relying purely on activation scale. To do this, we train a \textbf{No LN} transformer and at each step of training, we measure \emph{SR-train}: the train loss incurred when the scale of the MLP input is removed as $\text{MLP}(\bullet / \|\bullet\|)$. Note that we do not back-propagate the network on SR-train; it is measured purely to quantify sensitivity to scale. SR-train is found to be significantly larger than default train loss (with $\text{MLP}(\bullet)$), depicted in {Figure~\ref{fig:ib}a}. This implies strong sensitivity to feature scale instead of feature directions. Remarkably, SR-train reduces as the delayed descent in test loss, i.e., grokking, occurs and increases when the network memorizes at near-zero train loss and large test loss.
\paragraph{\cnum{2} $ \ $ Entropy of attention scores.} Embeddings as input to the query and key channel crucially make up the attention scores after projection by $W^Q$ and $W^V$. Here, the scale of these embeddings affects the sharpness of the entries of the softmax-ed product of queries and keys, thereby impacting the expressivity of the attention module. This is confirmed by measuring the average entropy of attention scores per row in configurations with and without LN on inputs to query and key channels of the attention module. As shown in {Figure~\ref{fig:ib}b} and {Figure~\ref{fig:ib}c}, the $\mathbf{A^{qk}}$ configuration\footnote{Superscript $\mathbf{qk}$ denotes LN is applied only on inputs to query and key channels of MHSA.} does not generalize within 10k epochs, falling behind \textbf{No LN} in both generalization and attention score entropy. Clearly, this shows that LN on inputs to query and key channels is detrimental to the degree of freedom of the network's MHSA. Additionally, this observation alings with mechanistic analyses of modular addition in transformers. Specifically,~\cite{nanda2023,clock2023} establish that the transformer's embedding table looks-up the tokens $a$ and $b$ to embeddings that denote $(\sin(\omega a), \cos(\omega a))$ and $(\sin(\omega b), \cos(\omega b))$. Then, the attention composes these embeddings via trigonometric identities in order to compute a representation proportional to $(\sin(\omega (a+b)), \cos(\omega(a+b)))$ at the last token position in the pre-logits. Specifically, attention makes the trigonometric compositions, while the MLP updates the hidden state of the last token. Under low entropy of the attention score softmax distribution, the network would limit the expression of the trigonometric identities that requires representations at both positions of $a$ and $b$, thereby hurting generalization in Figure~\ref{fig:ib}.
%
%
\paragraph{\cnum{3} $ \ $ Scale of value inputs.} The output of the attention module is linear in terms of the embeddings given as input to the value channel. This makes the result of adding the skip connection to the residual branch significantly dependent on embedding scale, which equivalently makes up the input to the MLP. Consequently, the value channel may also cause sensitivity to embedding scale in the MLP input pathway, as in \cnum{1}. To verify this claim, we introduce configuration $\mathbf{A^v}$ which only applies LN to the value inputs as $\text{MHSA}(x, x, \text{LN}(x))$ and train on our setting. From {Figure~\ref{fig:ib}b} and {Figure~\ref{fig:ib}c}, one can clearly see that LN only at the input of the value channel leads to a strong acceleration in grokking. However, $\mathbf{A^v}$ still groks slower than $\mathbf{M}$ because $\mathbf{M}$ fully removes scale from the MLP input. While $\mathbf{A^v}$ can limit radial variation contributed by the attention output, it does not do so to the extent as configuration $\mathbf{M}$. This is shown in {Figure~\ref{fig:ib}a} where $\mathbf{A^v}$ does not show increasing SR-train during the memorization period, i.e., before grokking, but also does not reduce SR-train strongly like $\mathbf{M}$. Overall, this demonstrates how the value channel of the attention can significantly impact grokking, and how sensitivity of MLP inputs to embedding scale is critical to fast generalization.
\section{Modulating \& Interpreting Grokking Through Optimization Choices}
\label{sec:lazy}
Having established how architectural design choices can affect inductive bias for grokking, we now move beyond considering only architecture as a source of inductive bias. Optimization choices such as learning rate and regularization are also a source of inductive bias, as they directly control the learning trajectory. For instance, higher learning rates can avoid certain solutions under high curvature, while high regularization may lead to slower fitting of the training data. Concretely, this section studies
\begin{tcolorbox}[before skip=8pt, after skip=8pt, boxrule=0.5pt, colback=gray!3]
\begin{center}
\emph{Beyond architecture, how can optimization choices affect inductive bias and rate of grokking?}
\end{center}
\end{tcolorbox}
To answer this question, we first recap the effects of learning rate and regularization in transformers learning to grok modular addition, as they have been studied extensively in the literature~\cite{liu2022towards,varma2023explaining}. Next, we consider another approach~\cite{lazy2rich} has controlled rate of grokking via readout scale, i.e., scaling the output of the network by some constant scalar $\alpha$. This study particularly uses readout scale $\alpha$ as a proxy for the network ``laziness'' in order to interpret grokking as a transition from lazy-to-rich regimes of feature learning. We borrow this approach and inspect its effects in our setting that importantly uses adaptive optimizers. Particularly, we relate its effects to optimization choices such as learning rate, regularization, and softmax-temperature, all of which guide the inductive bias of the network's learning.
\subsection{Preliminaries}
\begin{figure*}[t]
    \centering
    \includegraphics[width=\linewidth]{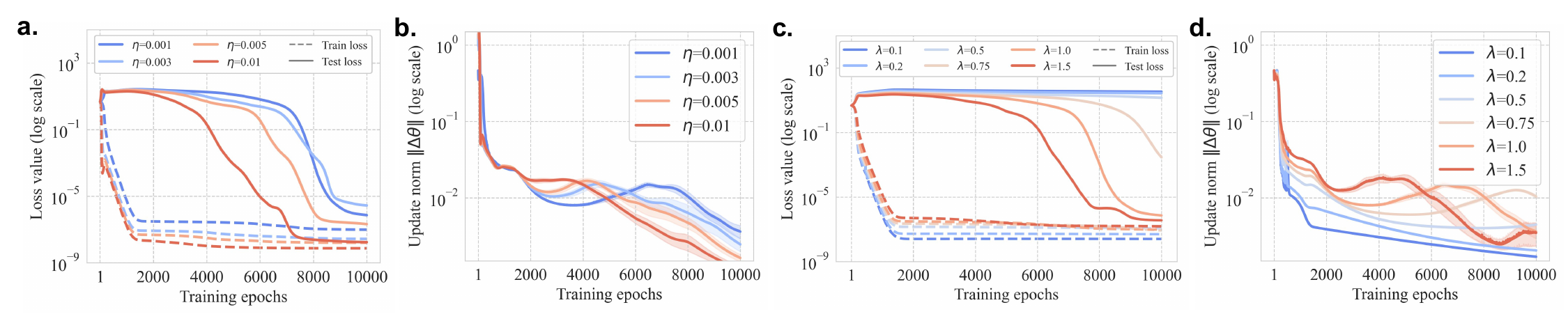}
    \caption{\textbf{(a)} Increasing learning rate $\eta$ while keeping weight decay constant leads to faster grokking. \textbf{(b)} Higher learning rates under fixed weight decay leads to significant differences in parameter update norm $\|\Delta \theta\|$ across training. \textbf{(c)} Similarly, increasing weight decay under constant learning rate leads faster grokking. \textbf{(d)} Higher weight decay strength also leads to large variations in parameter updates.}
    \label{fig:lr_wd_main}
\end{figure*}
\begin{figure*}[t]
    \centering
    \includegraphics[width=\linewidth]{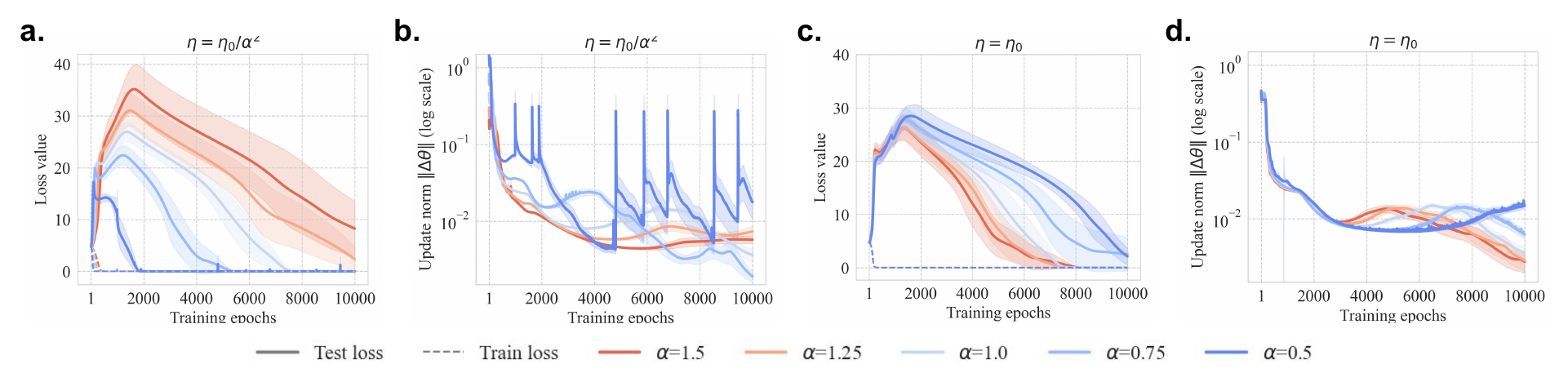}
    \caption{\textbf{(a)} Effect of varying learning rate as $\eta=\eta_0/\alpha^2$ on loss: higher readout scale $\alpha$ leads to slower grokking. \textbf{(b)} Parameter updates are not comparable across $\alpha$ under $\eta=\eta_0/\alpha^2$, and in fact diverge significantly. \textbf{(c)} Fixing learning rate to $\eta_0$ now leads to slower grokking with lower values of readout scale $\alpha$. \textbf{(d)} Constant learning rate $\eta_0$ lets parameter updates remain comparable across $\alpha$.}
    \label{fig:lazy_main}
\end{figure*}
Given a neural network $f(\bullet \ ; \ \theta)$ with parameters $\theta$,~\cite{lazy2rich} scale the network's output $f(x\ ; \theta)$ by a fixed scalar $\alpha > 0$ to get $h(x\ ; \theta) = \alpha f(x\ ; \theta)$ to control laziness, inspired from~\cite{chizat2019lazy}. Then, the gradient $g_{h} = \partial\mathcal{L}/\partial\theta$ of a loss function $\mathcal{L}$ for network $h$ would be
\begin{align}
    g_{h}&= \frac{\partial \mathcal{L}}{\partial h(x\ ; \theta)} \ \frac{\partial h(x\ ; \theta)}{\partial \theta} \\ &= \alpha \cdot \frac{\partial \mathcal{L}}{\partial h(x\ ; \theta)} \ \frac{\partial f(x\ ; \theta)}{\partial \theta}
    \label{eq:eq_1}
\end{align}
If $\mathcal{L}$ is the mean-squared error (MSE loss), i.e., for some input $x_i$ given to $h$ and ground truth $y_i$, $\mathcal{L}$ is given by
\begin{equation}
    \mathcal{L} = \frac{1}{2}\sum_i (h(x_i \ ; \theta) - y_i)^2.
    \label{eq:eq_2}
\end{equation}
Consequently, $\partial \mathcal{L}/ \partial h(x\ ; \theta) = \sum_i (\alpha f(x_i \ ; \theta) - y_i)$ which scales linearly with $\alpha$ and so
\begin{equation}
    g_h = \alpha \left( \sum_i (\alpha f(x_i \ ; \theta) - y_i) \right) (\partial f(x_i \ ; \theta)/\partial \theta) \sim \mathcal{O}(\alpha^2).
    \label{eq:eq_3}
\end{equation}
Thus the gradient and $\|\Delta \theta\| = \| -\eta \ g_h \|$ scales as $\alpha^2$ (ignoring weight decay). Then to keep weight update norms comparable in order to only change laziness and not effective step size, i.e., to ensure $\|\Delta \theta\| \approx \eta_0 \| g_f \|$ is fixed across $\alpha$, the learning rate must change as $\eta = \eta_0 / \alpha^2$. This learning rate scaling is followed in~\cite{lazy2rich} to present the lazy-to-rich interpretation of grokking. However, \emph{scaling the learning rate so is only applicable for MSE loss and SGD}. If we now consider $\mathcal{L}$ to be the cross-entropy loss
\begin{equation}
    \frac{\partial \mathcal{L}}{\partial h(x\ ; \theta)_j} = (p(\alpha f(x \ ; \theta))_j - y_j)
    \label{eq:eq_4}
\end{equation}
where $p(\bullet)$ is the softmax operation. Hence for cross-entropy loss, $\partial \mathcal{L}/\partial h(x\ ; \theta)$ does not scale linearly with $\alpha$. Hence, $g_h$ scales approximately linearly (certainly sub-quadratically) due to one factor of $\alpha$ coming from $\partial h(x \ ; \theta) / \partial \theta$. What is now crucial to note is that if the optimizer is chosen to be AdamW whose $\epsilon \to 0$, then we can write the weight update for the scaled network $h$ in terms of the Adam numerator $m_h$ and denominator $v_h$.
\begin{equation}
    \Delta \theta = -\eta \frac{m_h}{\sqrt{v_h} + \epsilon} \approx -\eta \frac{\alpha \cdot m_f}{\sqrt{\alpha^2 \cdot v_f}} = -\eta \frac{m_f}{\sqrt{v_f}}
    \label{eq:eq_5}
\end{equation}
\noindent This shows that parameter updates would be approximately comparable across $\alpha$, at least in the initial stages of training. The only differences in the updates would arise from the effect of $\alpha$ in $p(\bullet)$, however, the overall effect of $\alpha$ on the update would be approximately constant. To then choose $\eta = \eta_0 / \alpha^2$ would confound the comparison of laziness as the effective step size would be higher. This confounding effect can be understood as the optimizer governing inductive bias: using AdamW instead of SGD changes how readout scale affects the network.  
%
%
Also, PyTorch~\cite{paszke2019pytorch} and JAX~\cite{jax2018github} multiply weight decay strength with learning rate in their implementations of AdamW as shown in Equation~\ref{eq:wd_impl}. Therefore, using the $1/\alpha^2$ scaling would also confound regularization strength along with learning rate. Specifically for $\alpha < 1$ and $\alpha > 1$, the overall learning rate and weight decay would increase and decrease respectively. This will lead to different dynamics and the objective of keeping the parameter updates comparable across $\alpha$ would not be met.
\begin{equation}
\theta_{t} \leftarrow \theta_{t-1} - \eta \ \lambda \ \theta_{t-1}
    \label{eq:wd_impl}
\end{equation}
%
%
\subsection{Optimization choices modulate grokking}
\label{sec:validating_lazy_alpha}
We first observe the effects of optimization choices on grokking, by varying learning rate $\eta$ and weight decay $\lambda$. These are subsequently related with the effect of readout scale $\alpha$ in order to validate that readout scale can confound learning rate and weight decay under $\eta=\eta_0/\alpha^2$. Fixing $\eta=\eta_0$ shows that readout scale instead plays the role of inverse softmax-temperature, which also modulates grokking. These experiments clarify how grokking is shaped by inductive biases expressed through optimization choices by carefully relating existing interpretations with learning rate, weight decay, and softmax-temperature. Note that we only use the \textbf{No LN} configuration here. All other details remain the same as given in Appendix~\ref{sec:setup}.
%
%
%
\paragraph{Validating effects of learning rate and weight decay.} We show these effects of learning rate, keeping weight decay fixed regardless of $\alpha$, and vice-versa in {Figure~\ref{fig:lr_wd_main}}. As can be observed in {Figure~\ref{fig:lr_wd_main}a} and Figure~\ref{fig:lr_wd_main}b respectively, higher learning rates and weight decay strengths lead to faster generalization and lower grokking. Further, parameter update norms are not comparable in early stages of learning, visible in early epochs in {Figure~\ref{fig:lr_wd_main}b} and {Figure~\ref{fig:lr_wd_main}d}. Consequently, {Figure~\ref{fig:lazy_main}b} shows that scaling the learning rate as $\eta=\eta_0/\alpha^2$ leads to diverging and incomparable update norms, with grokking in {Figure~\ref{fig:lazy_main}a} correlate directly with $\alpha$ as $\alpha < 1 \implies$ large $\eta$ and vice-versa.
%
%
\paragraph{Enforcing comparable parameter updates.} To remove this discrepancy, incomparable updates afforded by scaling learning rate as $\eta=\eta_0/\alpha^2$, we keep effective step size a constant value $\eta_0=0.001$  across all $\alpha$. As visible in {Figure~\ref{fig:lazy_main}d}, update norms are comparable and well-behaved across all $\alpha$ in the initial stage of learning. Additionally, the trend visible in {Figure~\ref{fig:lazy_main}a} is the opposite of {Figure~\ref{fig:lazy_main}c} , i.e., with constant update norms, higher $\alpha$ generalizes faster. The reason for this can be attributed to effects of temperature in transformers and cross-entropy based tasks. The output scaling of $\alpha$ relates to softmax-temperature $\tau$ as $\alpha \propto 1/\tau$. Consequently, its effect as a proxy for $\tau$ modulates generalization~\citep{agarwala2020temperature} and grokking through its effect on optimization dynamics.
%
%
%
\\

In summary, using readout scale $\alpha$ as a control for laziness requires careful consideration of how the learning rate is scaled across different objective functions and optimizers. Crucially, scaling the learning rate as $\eta_0/\alpha^2$ leads to varying learning rate, weight decay, and update norms. We show how this confounds $\alpha$ as a control for laziness with regularization. Keeping the learning rate constant to allow for comparable parameter updates, we find that readout scale performs akin to softmax-temperature. Therefore, for our setting, readout scale to control laziness as per~\cite{lazy2rich} is in fact explained by how the learning process is biased through optimization settings. Consequently, characterizing the learning regime here becomes more nuanced than a lazy-to-rich transition.  
%
%
\section{Inductive Bias Pre-determines Evolution of Features in Grokking}
\label{sec:compressibility}
\begin{figure*}[t]
    \centering
    \includegraphics[width=\linewidth]{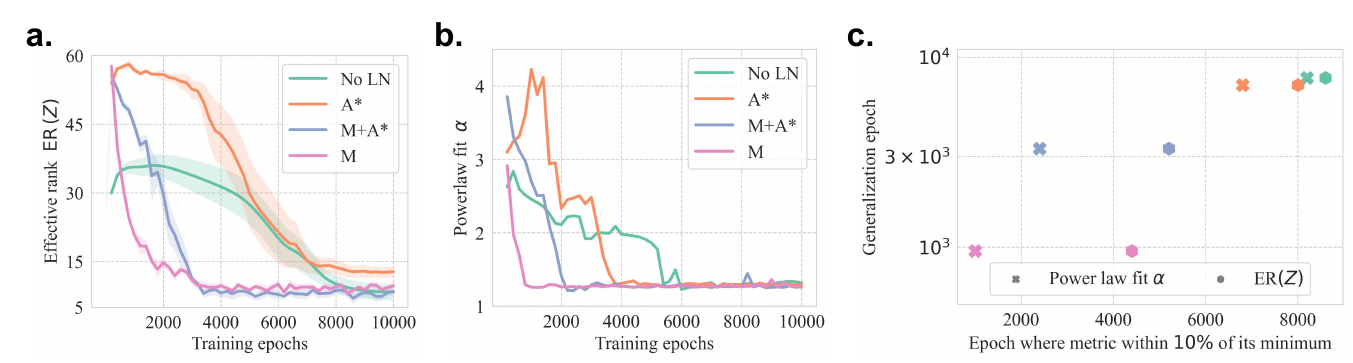}
    \caption{\textbf{(a)} Evolution of effective rank $\text{ER}(Z)$ of pre-logit correlation matrix $Z$ is shown averaged across $5$ seeds for each LN configuration. \textbf(b) Powerlaw fit $\alpha$ of the eigenspectrum of $Z$ is depicted across training. To eliminate noisy spikes caused by slingshots, this is shown under minimum aggregation over $5$ seeds. \textbf(c) The LN configuration that generalizes earlier arrives within $10\%$ of its respective minimum compressibility, measured as the average over $5$ seeds.}
    \label{fig:comp_ln}
\end{figure*}
\noindent Motivated by how a control for laziness can be explained through optimization-driven inductive biases, we now investigate the evolution of the network's features. 
This is because the nature of feature and parameter evolution can qualify the regime of feature learning that the network operates in.
Additionally, relating evolution of features to our findings on generalization across different modulators can reveal how inductive bias shapes the emergence of general features at large. Concretely, this section asks 
\begin{tcolorbox}[before skip=8pt, after skip=8pt, boxrule=0.5pt, colback=gray!3]
\begin{center}
\emph{What do features learned under different\\inductive biases move towards?} 
\end{center}
\end{tcolorbox}
Existing literature finds general features to exhibit specific compressibility trends~\cite{alphareq}. Consequently, we validate whether this is the case in our setting of transformers learning modular addition, by studying the evolution of feature compressibility.
\subsection{Relating Grokking to Feature Compressibility}
\noindent We first study compressibility of features learned via different LN positions in the transformer.
\paragraph{Experimental setup.} We use the same setup from {Section~\ref{sec:grokking_under_ln}}, however, we now study the effective rank of the correlation matrix $Z = (y_3+\widetilde{y}_3)^\top (y_3+\widetilde{y}_3)$ of the pre-logits of the ``='' token $(y_3+\widetilde{y}_3)$. This is denoted by ${\operatorname{ER}}(Z)$. We also want to quantify how aggressively compressibility is possible, i.e., if reconstruction error decays slowly with retained rank. For this, we fit a powerlaw to the eigenvalue spectrum of the feature correlation matrix. Coefficient of the powerlaw fit\footnote{We slightly abuse notation and use $\alpha$ to denote two things in our paper: readout scale in Section~\ref{sec:lazy} and powerlaw fit here.} $\alpha$ tells us how heavy-tailed the eigenspectrum is. Low $\alpha$ corresponds to more heavy tailed. Importantly, we only use data points from the training set and not the test set in order to check if training data is fit lazily. 
%
\paragraph{Features evolve significantly across training.} We begin by observing how the properties of features, i.e., pre-logits, evolve for the \textbf{No LN} configuration. Clearly, the correlation matrix $Z$ exhibits continuously changing effective rank as well as a variable eigenspectrum. This demonstrates significant changes to the features, corresponding to the rich regime of feature learning. 
\paragraph{Effect of LN position on compressibility.} Compressibility metrics largely follow the trend of grokking across LN positions. $\mathbf{M}$ shows the earliest onset of compressibility in the features, followed by $\mathbf{M}$+$\mathbf{A^{*}}$. Notably, $\mathbf{A^{*}}$ does not exhibit lower minimum effective rank than \textbf{No LN}, however it arrives within $10\%$ of its minimum effective rank slightly faster than \textbf{No LN}. This pattern aligns with how these configurations grok. {Figure~\ref{fig:comp_ln}a} and {Figure~\ref{fig:comp_ln}c} depict compressibility metrics averaged across $5$ seeds. When observing powerlaw fit $\alpha$ in {Figure~\ref{fig:comp_ln}b}, we plot a min aggregation of $5$ seeds to eliminate spikes caused by slingshots, and find that the order of grokking is aligned here. Specifically, networks that grok faster arrive at low values show more heavy-tailed eigenspectra in $Z$. 
\paragraph{Compressibility is aligned across modulators.} We relate the compressibilities of features under different LN positions with learned under of different weight decay strengths in transformers without LN. Higher weight decay, shown to induce less grokking and faster generalization in {Section~\ref{sec:validating_lazy_alpha}}, leads to more compressibility in $Z$ via reduction in effective rank and powerlaw fit $\alpha$. This is shown in {Figure~\ref{fig:wd_comp}} of {Appendix~\ref{sec:wd_comp}}. Additionally, higher values of weight decay show more heavy-tailed eigenspectra in $Z$. These effects of weight decay, across test loss and compressibility, are quite similar those of LN position: configurations that generalize earlier arrive at greater compressibilities earlier.
\subsection{Discussing LN's effect on inductive bias, compressibility, \& generalization}
\label{sec:com_disc}
Here, we highlight that the onset of compressibility for different LN positions is strongly correlated with generalization trends exhibited in Section~\ref{sec:grokking_under_ln}. Given that~\cite{alphareq} link powerlaw fit $\alpha$ to generalization of the network, it is notable to relate this to the inductive biases studied in Section~\ref{sec:ln_ib}. Specifically, configurations sensitive to scale of the MLP's inputs generalize slower, and exhibit higher $\alpha$ in Figure~\ref{fig:comp_ln}. We link this finding to existing work on memorization in transformer MLPs~\cite{mlpmem}. Relevantly,~\cite{stephenson2021geometry} link memorization to feature geometry, i.e., manifold radius and dimension. 
%
\section{Discussion, Conclusion, \& Limitations}
\label{sec:conclusion}
\paragraph{Connections to previous work.} By showing the variation in inductive bias afforded by different optimization schemes, our work suggests against a lazy-to-rich viewpoint~\cite{lazy2rich} in transformers when learning modular addition. In a similar vein, yet not precisely for grokking,~\cite{chou2025feature} show how learning features can be more nuanced that a dichotomous lazy-versus-rich characterization by studying the separation of task-relevant manifolds. Specific to transformers, our study on how biases in particular pathways strongly relate to the network's generalization aligns with existing work, seen through the lens of compressibility in Section~\ref{sec:com_disc}. Appendix~\ref{sec:litreview} expands such connections further. 
\paragraph{Conclusion.} Our work shows how architectural modifications on specific pathways predisposes shortcut-learning and attention entropy in the network. Additionally, biases from different optimization settings lead to different interpretations of the readout scale, which we study through modulating learning rate, weight decay, and softmax-temperature. These findings comprehensively relate inductive biases expressed through architecture and optimization to grokking behavior. Finally, we show how features of all data points change significantly and continuously throughout the learning process. This evolution sees grokking as more than a transition from lazily fitting the training data to learning rich general features. Our findings suggest that the network learns features richly throughout. Additionally, existing studies on generalizable features~\cite{alphareq} afford a setting-agnostic progress measure for grokking, which aligns with how generalization emerges with feature evolution predictably across different modulators of inductive bias in out work.
\paragraph{Limitations and future work.} This work is particularly restricted to transformers learning modular addition under adaptive optimizers. Accordingly, our findings on specific inductive biases, their role in grokking, and relations to lazy-versus-rich learning may not apply to other settings. Instead, our focused scope links to a broader interest in layer normalization~\cite{ba2016layernormalization} for architecture design and its role in shaping the optimization trajectory, by presenting a toy model of how it may impact learning. However, we note that this work is not an exhaustive study of all possible LN positions. Notably, while our main manuscript focuses on one-layer transformers, our findings also hold for transformers of depth 3 under the same setup, depicted in Appendix~\ref{sec:more_layers} Finally, using of~\cite{alphareq} for grokking in Section~\ref{sec:compressibility} brings out a progress measure for grokking that is plausibly setting-agnostic. Future work may utilize this measure for grokking setups at large.
\section{Impact Statement}
This work advances understanding of delayed generalization in neural networks by analyzing grokking through controlling inductive biases in small transformers solving a toy task. It is focused on clarifying existing interpretations of the grokking phenomenon and does not introduce new models, datasets, or applications intended for deployment in real-world systems. As such, we do not foresee direct positive or negative societal impacts arising from this work beyond its contribution to scientific understanding.
\section{Acknowledgements}
The authors thank the International Max Planck Research School for Intelligent Systems (IMPRS-IS) for supporting Diganta Misra. Jaisidh Singh is supported by the Konrad Zuse School of Excellence in Learning and Intelligent Systems (ELIZA) through the DAAD programme Konrad Zuse Schools of Excellence in Artificial Intelligence, sponsored by the Federal Ministry of Education and Research. This work was enabled by compute resources provided by Max Planck Institute for Intelligent Systems Tübingen \& Amazon Science Hub.
%
\bibliographystyle{icml2026}
\def\UrlBreaks{\do\/\do-\do_}
\bibliography{main}
\newpage
\onecolumn
%
%
\clearpage
\appendix
\part*{Appendix}
\section{Related Work}
\label{sec:litreview}
\paragraph{Feature learning and kernel dynamics.} Feature learning, or the ability of a model to change its activations in order to solve a given task has been a core area of interest in machine learning research. Works like \cite{chizat2019lazy} show that it is possible to perform well on tasks without changing the network's parameters and features, i.e, learning can occur in the lazy regime. Moreover, initial weight norm, readout scale, and learning rate were found to be among the factors that can govern whether a network operates in the lazy or rich regime of feature learning. The seminal work of Neural Tangent Kernel (NTK) \cite{ntk} has been integral in characterizing the behavior of neural networks, and becomes relevant to feature learning as in the lazy regime, the network drifts negligibly from its NTK at initialization. In other words, the network stays in its linearized form as admitted by the NTK at initialization. Notably, the NTK can still be defined even when the network operates in the rich regime, however, drift from NTK at initialization would be significant in this case. Consequently, this line of research has motivates other studies which investigate distinctions and overlaps between the conditions inducing these regimes. For instance,  \cite{tp4,shi2022theoretical,karp2021local,damian2022neural} have shown that conditions inducing lazy learning can also admit rich feature learning, and that the rich regime can allow networks to perform better than when they learn lazily.
\paragraph{Inductive bias, memorization, and generalization.} Learning is typically a search over a solution space, to find solutions that can best fit or explain the data distribution. It may be possible that for learners such as neural networks, many solutions may appropriately fit the data~\cite{goodman1965new}. Moreover, since unseen situations can have arbitrary output values, one requires a set of assumptions that constrain the learning process to particular problem-specific behaviors. These assumptions paired with the constrained tendencies of the learner are what machine learning practitioners term as inductive bias~\cite{gordon1995evaluation}. Accordingly, inductive bias is responsible for the model prioritizing one pattern of solving the problem over another~\cite{mitchell1980need}. This makes inductive bias directly relevant to the behavior of the model w.r.t. generalization and memorization. Generalization is perhaps the main goal of machine learning methods, and has been studied across many interpretations and settings~\cite{zhang2016understanding,ramesh2023compositional}. Crucially, generalization as well as memorization has been found to be affected by various properties of features. For instance,~\cite{advani2020high} find that eigenvalues of input covariance determine
the generalization error dynamics. On the other hand, deep sequence models are shown to have puzzling tendencies to memorize geometrically~\cite{noroozizadeh2025deep}, and that MLP blocks in transformers are responsible for the bulk of the memorization~\cite{mlpmem}.
%
%
\paragraph{Grokking and modular addition.} Grokking has emerged as a bridge between mechanistic interpretability~\cite{olah2020zoom,iclandinductionheads,wang2022interpretability,knowledgecircuits} and the study of training dynamics~\cite{liu2022towards,liu2022omnigrok,davies2023unifying}. The setting we study, i.e., the popular modular addition task~\cite{gromov2023grokking} ($a+b\mod p$) learned by transformers was first mechanistically analyzed by~\citep{nanda2023} to reveal that generalizing solutions contained periodic structure in the parameter matrices of the transformer, such as the embedding matrix and the neuron-logit map. Input tokens that require addition, i.e, $a$ and $b$  are mapped to rotations $(\sin(\omega a), \cos(\omega a))$ and $(\sin(\omega b), \cos(\omega b))$ respectively by the embedding matrix. These rotations are then composed by the transformer using trigonometric identities to then depict $(\sin(\omega(a+b)), \cos(\omega(a+b))$ in the feature of the last token. This feature is then unembedded to predict the current result of the modulus. Interestingly,~\cite{clock2023} find that this solution is not unique, and show that transformers can learn the ``pizza'' algorithm in addition to several other solutions that are related, yet different from that found by~\citep{nanda2023}, (the ``clock'' algorithm). Moreover, transformers can exhibit sharp phase transitions~\cite{akyurek2022learning} between these two solutions depending on width and attention strength. These results arise under specific settings of optimization and data starvation, discussed in more detail in~\citep{liu2022towards}. Specific to feature learning and lazy-versus-rich characterizations,~\citep{lazy2rich} interpret grokking as a transition from a lazy, kernel-like regime to a rich regime of feature learning. Using readout scale as a proxy for laziness, they show higher readout scales delay generalization in several settings. However, for transformers solving modular addition, varying learning rate with readout scale $\alpha$ as done in~\citep{chizat2019lazy} does not allow for comparable update dynamics across $\alpha$ in adaptive optimizers. We study this in \emph{Section~\ref{sec:lazy}}, showing that effects of $\alpha$ can be confounded with those of hyper-parameters, hence showing limited applicability of the lazy-to-rich interpretation. 
\section{Notation}
\label{sec:notation}
\begin{table}[h]
    \centering
    \begin{tabular}{l|l|l}
        \toprule
        \textbf{Notation} & \textbf{Dimension / Mapping} & \textbf{Description} \\
        \midrule
        $W_E$ & $\mathbb{R}^{d\times (p+1)}$ & Embedding table of transformer \\
        $\rho$ & $\mathbb{R}^{3\times d}$ &Position encoding \\
        $\operatorname{MHSA}(\bullet)$ & $\mathbb{R}^{d}\to\mathbb{R}^d$ & Multi-head self attention module \\
        $W^Q$ & $\mathbb{R}^{d\times (d/h) \times h}$ & Query weight matrix of attention module \\
        $W^K$ & $\mathbb{R}^{d\times (d/h) \times h}$ & Key weight matrix of attention module \\
        $W^V$ & $\mathbb{R}^{d\times (d/h) \times h}$ & Value weight matrix of attention module \\
        $\operatorname{qkv\_prod}(\bullet)$ & $\mathbb{R}^{d} \to \mathbb{R}^d$ & Multi-head dot product of query, key, and value vectors. \\
        $W^O$ & $\mathbb{R}^{d\times d}$ & Out projection of attention module \\
        $\operatorname{MLP}(\bullet)$ & $\mathbb{R}^d \to \mathbb{R}^d$ & Multi-layer perceptron module of transformer\\
        $W_1$ & $\mathbb{R}^{D \times d}$ & Input projection of MLP\\
        $\sigma(\bullet)$ & $\mathbb{R}^{D} \to \mathbb{R}^{D}$ (element-wise) & non-linear activation function (ReLU)\\
        $W_2$ & $\mathbb{R}^{d \times D}$ & Output projection of MLP\\
        $W_U$ & $\mathbb{R}^{(p+1) \times d}$ & Un-embed matrix of transformer\\
        \bottomrule
    \end{tabular}
    \caption{We describe the notation used in our transformer above.}
    \label{tab:notation_table}
\end{table}
\section{Experimental Setup} 
\label{sec:setup}
We follow~\cite{nanda2023} and use a one-layer transformer of width 128, multi-head self-attention with 4 heads, a hidden dimension of 512 in the transformer's MLP, no weight tying in the unembed. The configurations given in Section~\ref{sec:grokking_under_ln} are used to create variants of the transformer defined by positions of LN. Note that in Section~\ref{sec:lazy}, there is no LN anywhere in the model. In our experiments, transformer configurations containing LN will be explicitly mentioned and displayed in figures. If not, then it indicates that LN is absent in the transformer. Optimization is always performed full batch for $10000$ epochs using AdamW with learning rate $0.001$, $(\beta_1, \beta_2)=(0.9, 0.98)$, unscaled weight decay of $\lambda=1.0$, and $\epsilon=1e-8$. 
The train-test data split is $30$-$70\%$ with the total number of samples being $p^2$. The network is initialized with a truncated normal initialization similar to \texttt{HookedTransformer} from \texttt{TransformerLens}~\cite{nanda2022transformerlens}. All experiments are implemented in PyTorch~\cite{paszke2019pytorch} and are executed on a single GPU. Recall that in PyTorch and JAX~\cite{jax2018github}, weight decay applied in the update is scaled by the learning rate, indicated in Equation~\ref{eq:wd_impl}.
\section{Implementation of Metrics}
\label{sec:impl}
\subsection{$\mathcal{C}_{\text{Fourier}}$}
\begin{lstlisting}
P = 113

def variance_ratio(X):
    row_means = X.mean(axis=1, keepdims=True)
    between = np.var(row_means)
    within = np.mean((X - row_means)**2)
    return between / (within + 1e-8)

W_e = model.embed.weight.data.clone()[:P, :]
W_f = torch.fft.rfft(w_e, dim=0).abs().cpu().numpy()
c_fourier = variance_ratio(W_f)
\end{lstlisting}
\newpage\subsection{ER$(Z)$}
\begin{lstlisting}
def compute_esd(X):
    n, m = X.shape
    if n != m:
        eigenvalues = torch.linalg.svdvals(X)
    else:
        eigvals = torch.linalg.eigvalsh(X @ X.T)
        eigenvalues = torch.sqrt(torch.abs(eigvals))
    return torch.sort(eigenvalues, descending=True).values

Z = prelogits[:, -1, :].T @ prelogits[:, -1, :]
ev = compute_esd(Z)
p = ev / torch.linalg.norm(ev, ord=1)
entropy = -1 * (p * torch.log(p)).sum()
effect_rank = torch.exp(entropy).item()
\end{lstlisting}
\subsection{Powerlaw fit $\alpha$}
\nocite{alstott2014powerlaw}
\begin{lstlisting}
from powerlaw import Fit 
# the `powerlaw` package (Alstott el al., 2014) 
# can be installed via `pip install powerlaw`

def compute_esd(X):
    n, m = X.shape
    if n != m:
        eigenvalues = torch.linalg.svdvals(X)
    else:
        eigvals = torch.linalg.eigvalsh(X @ X.T)
        eigenvalues = torch.sqrt(torch.abs(eigvals))
    return torch.sort(eigenvalues, descending=True).values

def esd_powerlaw(X):
    evs = compute_esd(X)
    eigenvalues_np = evs.cpu().numpy()
    fit = Fit(eigenvalues_np, discrete=False, verbose=False)
    alpha = fit.alpha
    return alpha

Z = prelogits[:, -1, :].T @ prelogits[:, -1, :]
alpha = esd_powerlaw(Z)
\end{lstlisting}
\section{Using RMSNorm in place of LayerNorm}
\label{sec:rmsnorm}
\begin{figure}[!h]
    \centering
    \includegraphics[width=0.4\linewidth]{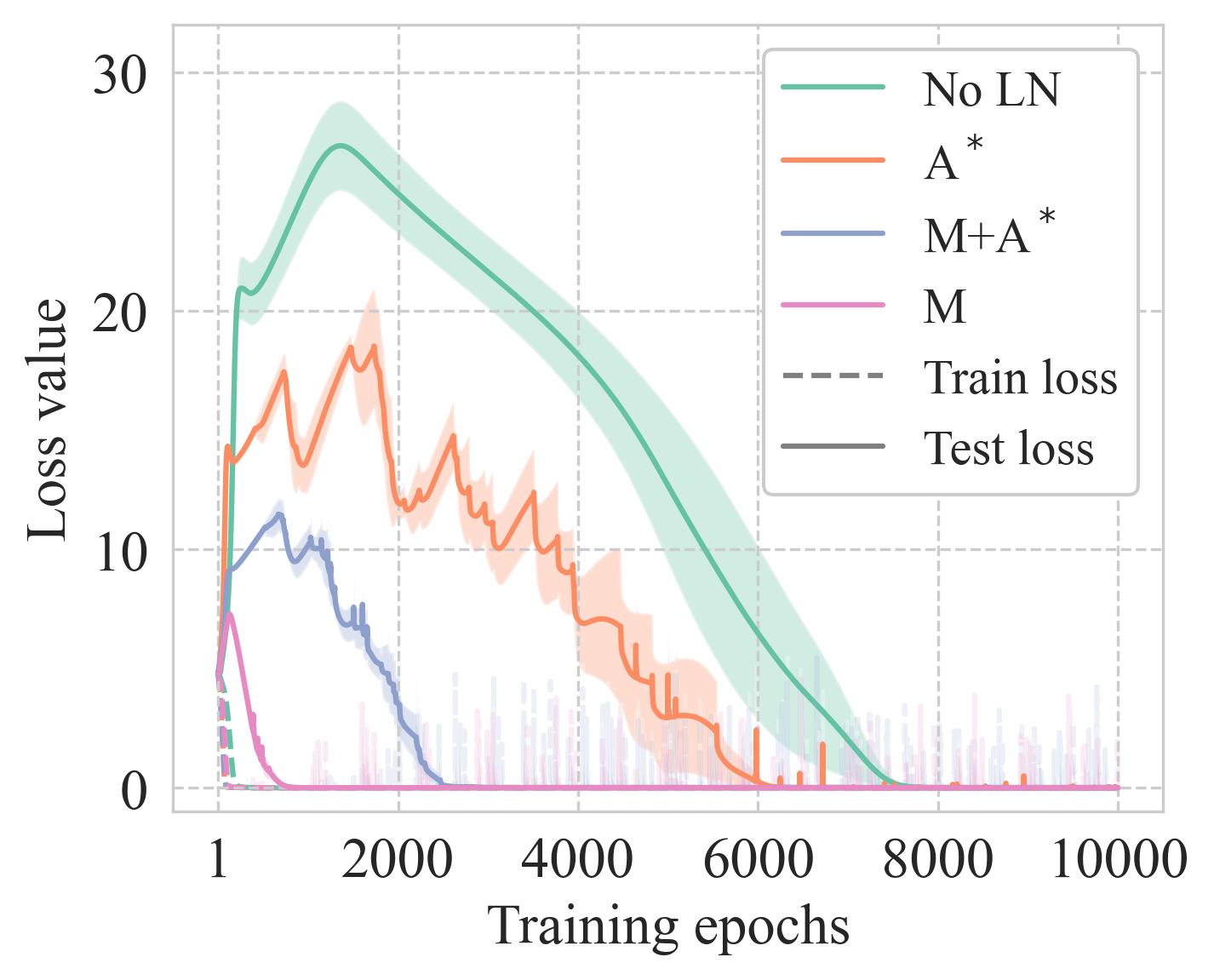}
    \caption{Using RMSNorm in the transformer shows very similar results.}
    \label{fig:rmsnorm}
\end{figure}
\section{Effect of LN position on deeper transformers}
\label{sec:more_layers}
\begin{figure}[!h]
    \centering
    \includegraphics[width=0.4\linewidth]{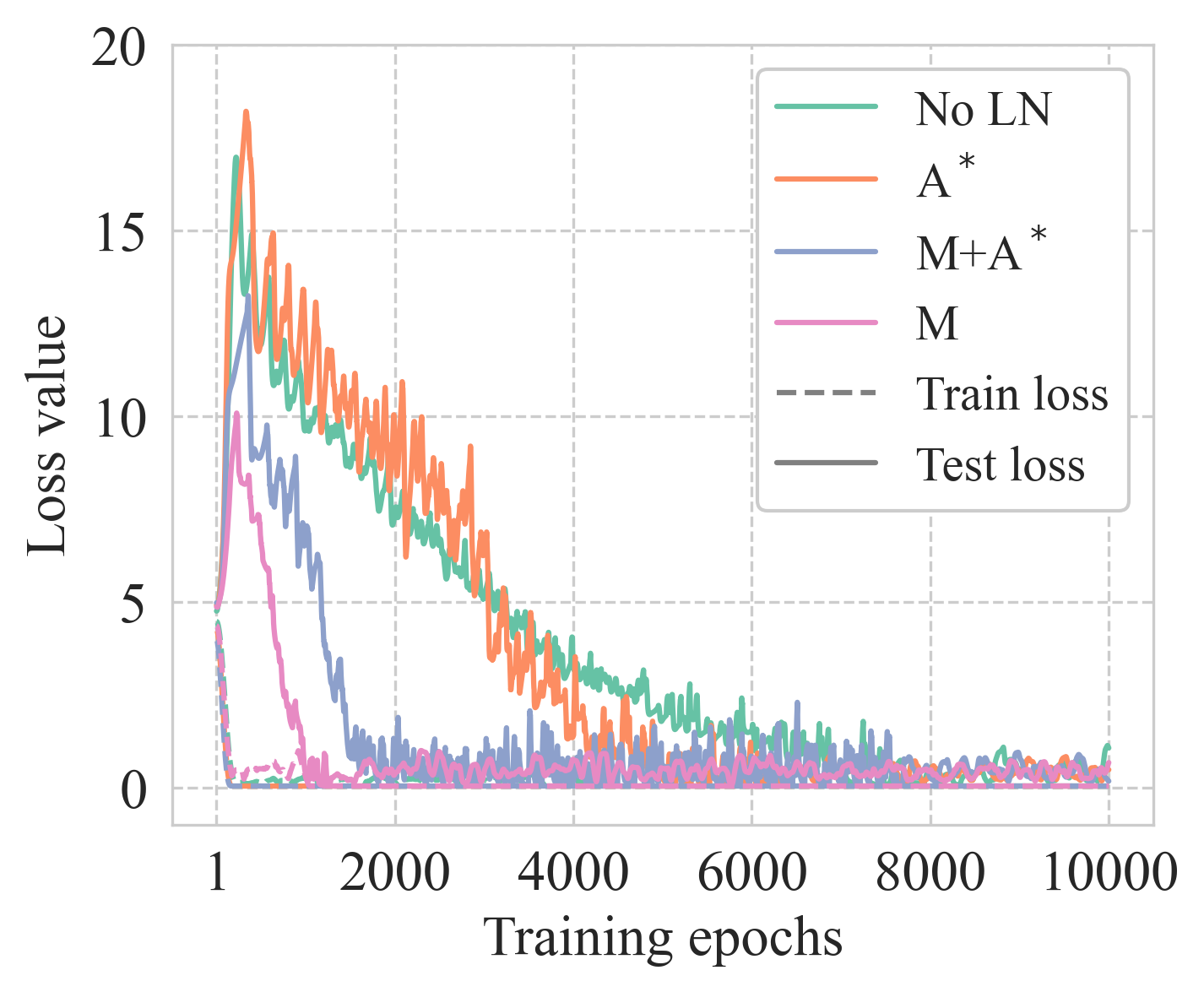}
    \caption{We increase depth of the transformers to 3 layers. Width is reduced to 64 to limit capacity of the network to see grokking. Loss curves are smoothed as more depth shows significantly more slingshots~\cite{slingshot}, an observation that is consistent with the findings of~\cite{nanda2023}. Overall, grokking behavior under this setup is very similar to that of one-layer transformers.}
    \label{fig:d3_w64}
\end{figure}
\section{Effect of weight decay on compressibility}
\label{sec:wd_comp}
\begin{figure}[!h]
    \centering
    %
    \begin{subfigure}{0.45\linewidth}
        \includegraphics[width=\linewidth]{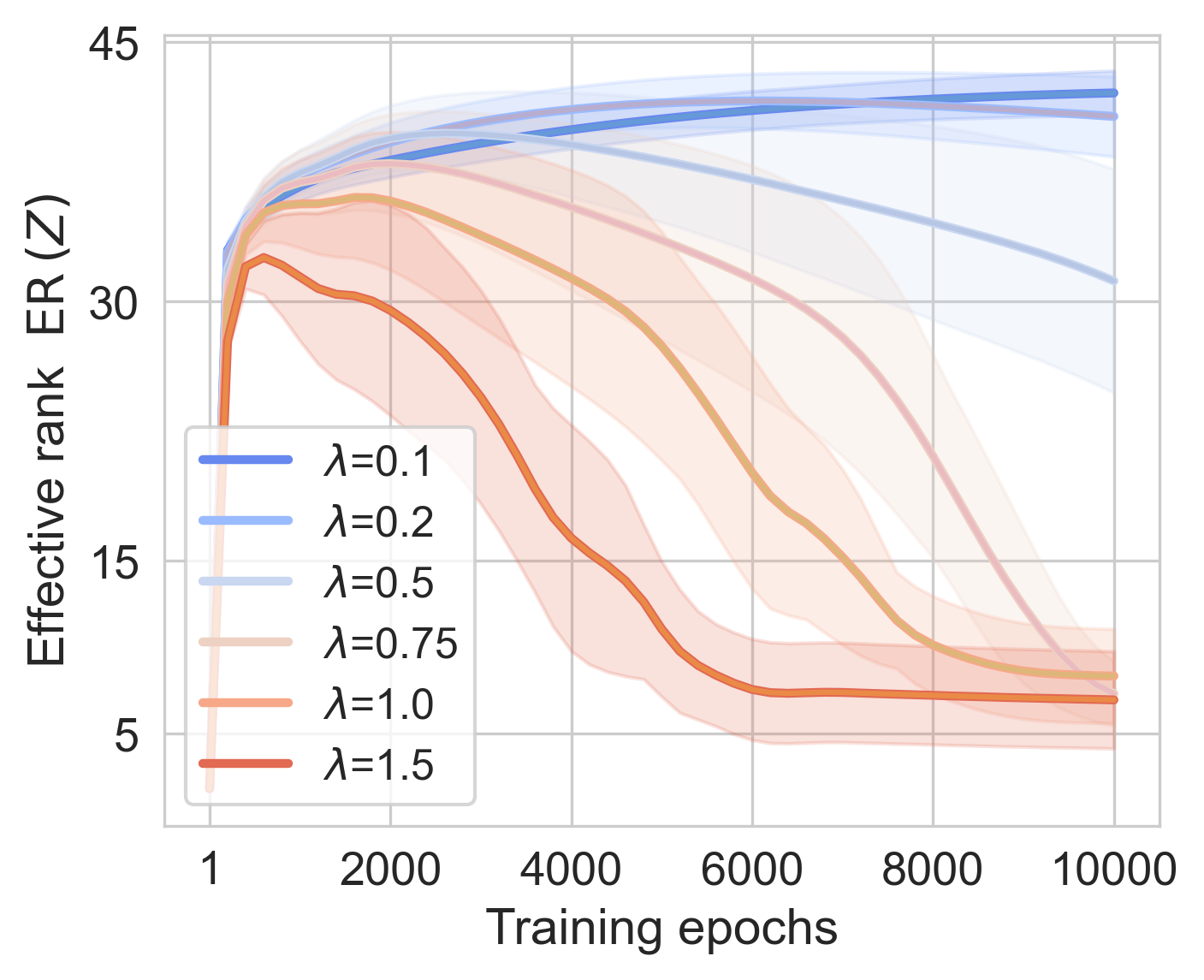}
        \caption{Effect on $\lambda$ on $\operatorname{ER}(Z)$}
    \end{subfigure}
    \begin{subfigure}{0.45\linewidth}
        \includegraphics[width=\linewidth]{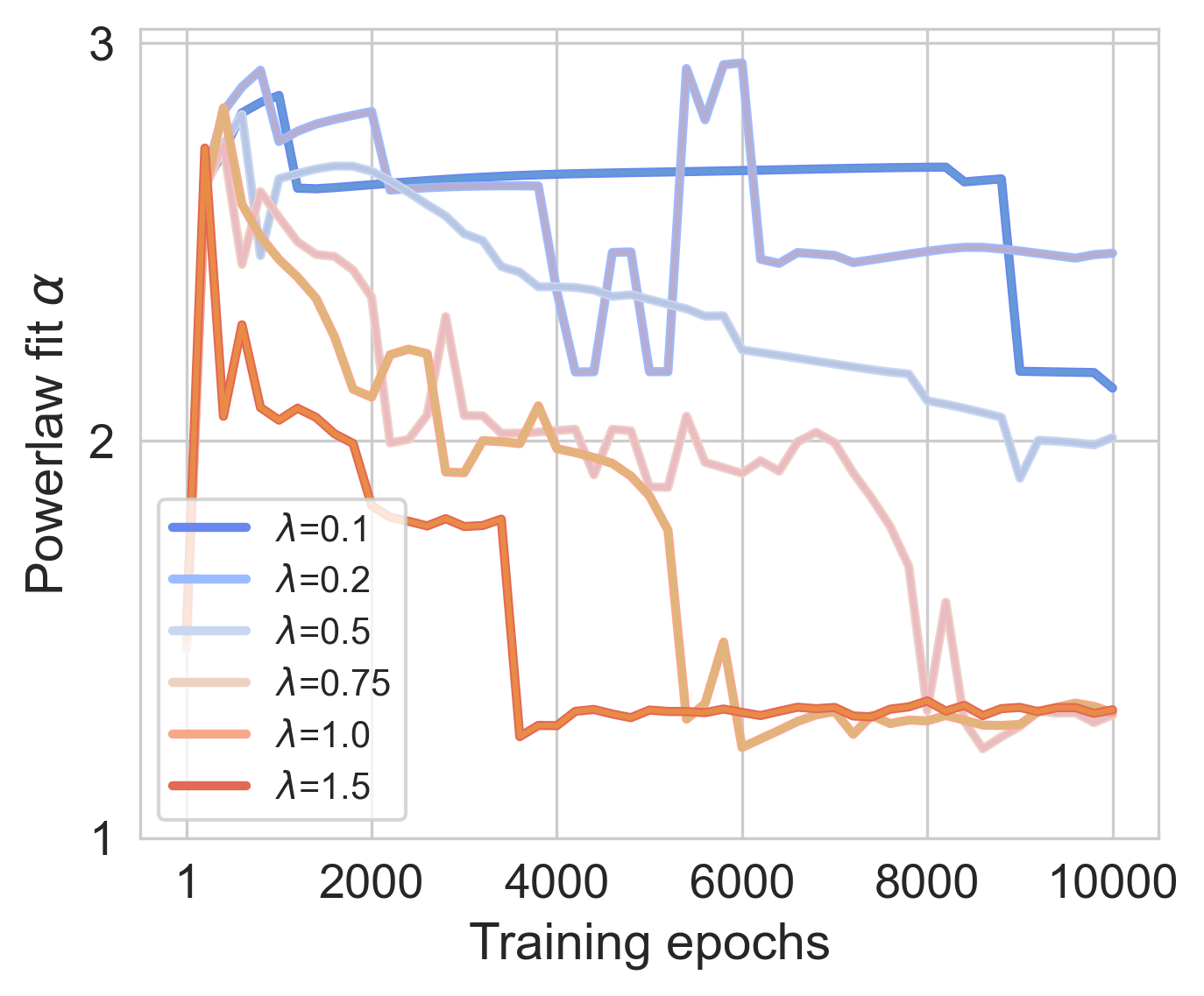}
        \caption{Effect of $\lambda$ position on powerlaw fit $\alpha$}
    \end{subfigure}
    %
    \caption{Effects of weight decay $\lambda$ on compressibility metrics.}
    \label{fig:wd_comp}
\end{figure}
%
%
\end{document}